\documentclass[conference,a4paper]{IEEEtran}

\usepackage{graphicx}
\usepackage{subfig}
\usepackage[absolute,showboxes]{textpos}

\TPMargin{5pt}

\newcommand{\copyrightstatement}{
    \begin{textblock}{0.85}(0.08,0.90)    
         \noindent
         \footnotesize
Copyright 2018 IEEE. Published in the Digital Image Computing: Techniques and Applications, 2018 (DICTA 2018), 10-13 December 2018 in Canberra, Australia. Personal use of this material is permitted. However, permission to reprint/republish this material for advertising or promotional purposes or for creating new collective works for resale or redistribution to servers or lists, or to reuse any copyrighted component of this work in other works, must be obtained from the IEEE. Contact: Manager, Copyrights and Permissions / IEEE Service Center / 445 Hoes Lane / P.O. Box 1331 / Piscataway, NJ 08855-1331, USA. Telephone: + Intl. 908-562-3966.
    \end{textblock}
}

\begin{document}
%
\title{{\em In Situ} Cane Toad Recognition}


\author{\IEEEauthorblockN{
Dmitry A. Konovalov}
\IEEEauthorblockA{College of Science and Engineering\\
James Cook University\\
Townsville, Australia\\
dmitry.konovalov@jcu.edu.au}
\and
\IEEEauthorblockN{Simindokht Jahangard}
\IEEEauthorblockA{Medical Image and \\
Signal Processing Research Center\\
Isfahan University of Medical Science\\
Isfahan, Iran\\
s.jahangard66@gmail.com}
\and
\IEEEauthorblockN{Lin Schwarzkopf}
\IEEEauthorblockA{College of Science and Engineering\\
James Cook University\\
Townsville, Australia\\
lin.schwarzkopf@jcu.edu.au}
}



\maketitle

\copyrightstatement

\begin{abstract}
\boldmath
Cane toads are invasive, toxic to native predators, compete with native insectivores, and have a devastating impact on Australian ecosystems, 
prompting the Australian government to list toads as a key threatening process under the Environment  Protection  and Biodiversity Conservation Act 1999. 
Mechanical cane toad traps could be made more native-fauna friendly if they could distinguish invasive cane toads from native species. Here we designed 
and trained a Convolution Neural Network (CNN) starting from the Xception CNN. 
The XToadGmp toad-recognition CNN we developed was trained end-to-end 
using heat-map Gaussian targets. After training, XToadGmp required minimum 
image pre/post-processing and when tested on 720x1280 shaped images,
it achieved 97.1\% classification accuracy on 1863 toad and 2892 not-toad test images, which were not used in training.

\end{abstract}


%
\IEEEpeerreviewmaketitle

\section{Introduction}

In Australia, the cane toad ({\em Rhinella marina}, formerly {\em Bufo marinus}) 
is an invasive pest species. Native to Central and South America, the toads were deliberately released in the Australian state of Queensland in 1935 in an attempt to control pests of sugar cane, including the beetle ({\em Dermolepida albohirtum}).
Because cane toads are invasive  \cite{Rapid2007}, 
toxic to some native predators, compete with native wildlife, and they can have a devastating impact on Australia's ecosystems, the Australian government has listed cane toads  as a key threatening process under the Environment Protection and Biodiversity Conservation Act 1999
\cite{ToadPlan2005,ToadPlan2011}.   

One approach to controlling invasive cane toads is to deploy mechanical traps 
\cite{ToadTrap2018}, 
which use a lure and a cane toad vocalization to attract and trap adult toads. 
An LED ultraviolet light is also used to attract insects to the vicinity of the trap, which further enhances the trap attractiveness. Adult cane toads are nocturnal 
\cite{ToadTookit2017} 
and therefore the mechanical traps are most effective at night when at least some Australian native wildlife are also active. Trapping bycatch is a highly undesirable consequence of {\em blind} mechanical traps, which by their design are not able to distinguish among wildlife types (e.g., desirable catch versus bycatch). This study reports the first step in developing a computer vision system to recognize cane toads in traps in the field.
If this approach is successful, it may be possible to modify traps to be selective.


The field of computer vision is currently dominated by Deep Learning Convolution Neural Networks (CNNs) \cite{LeCun15}. 
A large variety of classification CNNs are now readily and freely available for download \cite{keras}. 
A typical {\em off-the-shelf} CNN was trained to recognize 1,000
object classes from the Image Net collection of images \cite{ImageNet2012}. 
Some popular CNNs such as ResNet50 \cite{ResNet}, InceptionV3 \cite{InceptionV3} and Xception \cite{Xception} have arguably reached an accuracy saturation level for practical application, where they achieved similar state-of-the-art classification accuracy \cite{NasNet2017}.
Furthermore, the Image-Net-trained CNNs are often more accurate than randomly initialized CNNs (of the same architecture), when they are re-purposed for other object classes \cite{Oquab_2014_CVPR}.
This effect is known as the {\em knowledge transfer} \cite{Oquab_2014_CVPR} property of the Image-Net-trained CNNs. 
The ability to re-train and easily re-purpose existing Image-Net-trained CNNs was considered essential for this study.
For that reason, the user-friendly high-level neural network Application-Programming-Interface Keras \cite{keras} was 
used in this study together with the machine-learning Python package TensorFlow \cite{tensorflow2015-whitepaper}. 

Working with actual in-situ video clips, 
in this paper we developed a novel approach of training classification CNNs by 
manually and approximately segmented target binary masks. 
When the masks were converted to Gaussian heat-maps, a fully convolutional CNN  
was successfully trained by the Mean Squared Error loss function on the highly imbalanced training dataset (90\% negative and 10\% positive toad-containing images). Once trained the XToadHm CNN was
converted to the final toad/not-toad classifier (XToadGmp) by adding a single spatial maximum pooling layer.
The final XToadGmp classifier was tested on holdout video frames, which were not used in training, and achieved a classification accuracy of 97.1\%, sufficient for practical real-time detection. 
Furthermore, and most encouragingly, XToadGmp delivered 0\% false-positive misclassifications thereby fulfilling its main ecological goal of not confusing native species with invasive cane toads. 
Our approach demonstrated that only 66 toad bounding rectangular-boxes were sufficient to train the very accurate toad/not-toad XToadGmp detector.
This work confirmed the suitability of the rectangular training masks, which could be obtained manually or by other CNNs 
for a much larger number of training images in the future.

The structure of this paper is as follows. Section~\ref{subsec:dataset} describes 
the images extracted from {\em in-situ} video clips. 
Section~\ref{subsec:heatmap} explains how the Xception CNN was used to create XToadHm CNN, 
which could be trained by manually segmented toad binary masks. 
Section~\ref{section:training} presents the training pipeline using extensive image augmentation steps.   
Section~\ref{subsec:gauss} introduces the main novel aspect of this work: 
training classification CNN by Gaussian heat-maps.
Section~\ref{sec:results} presents the achieved results on the test images not used in training of the XToadHm/Gmp CNNs.

\begin{figure}[!ht]
\begin{center}
\subfloat[Cane toad on a plain background. \label{subfig:toads-a}]{%
\includegraphics[width=0.35\textwidth]{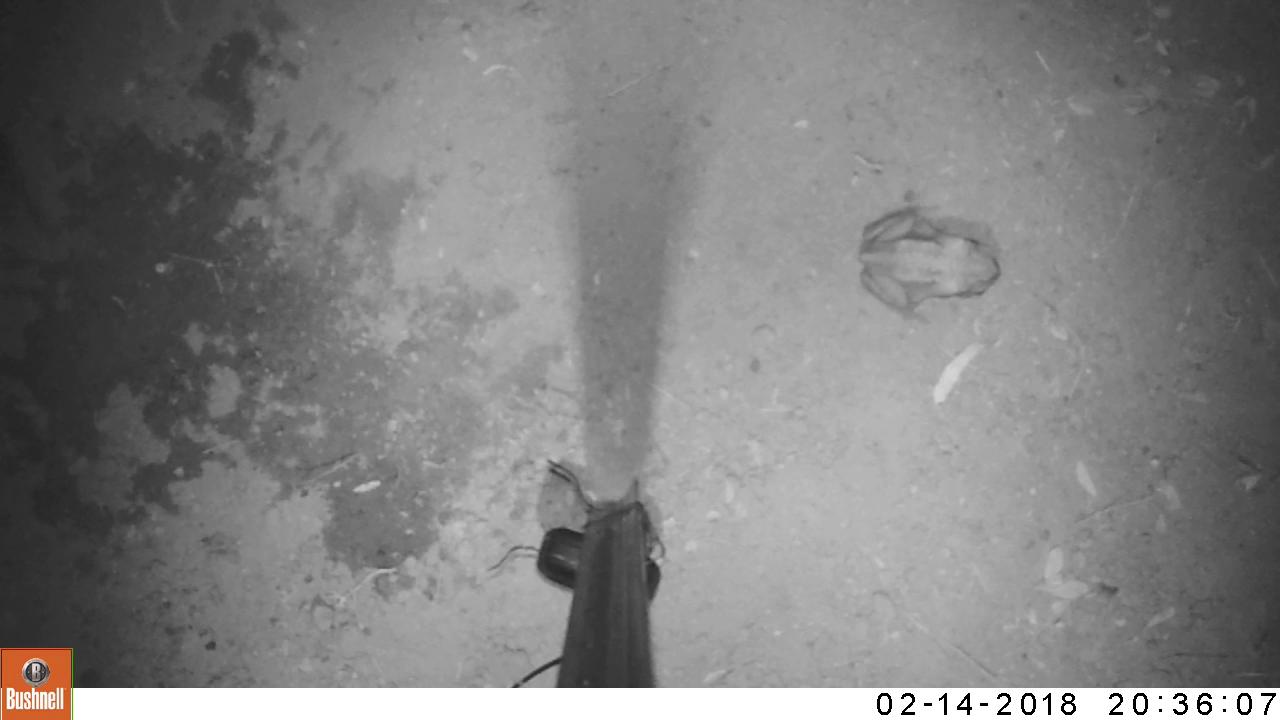}
}
\hfill
\subfloat[Two cane toads on a complex background. \label{subfig:toads-b}]{%
\includegraphics[width=0.35\textwidth]{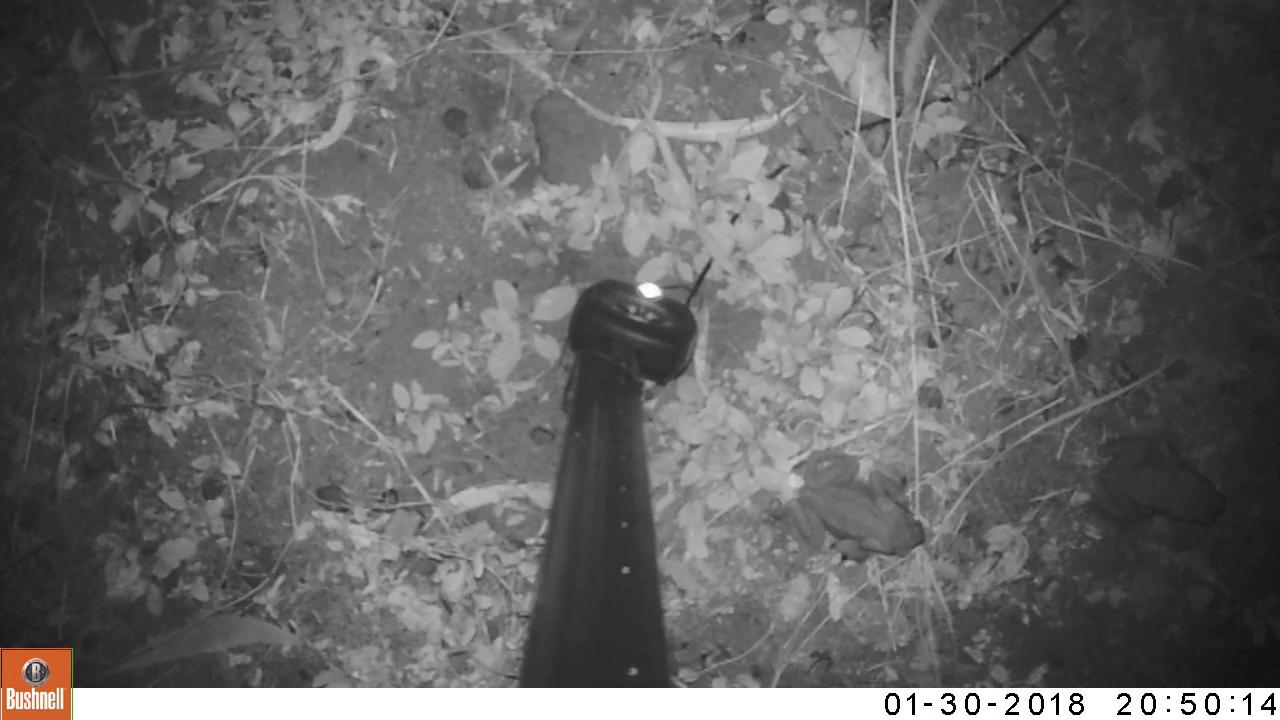}
}
\hfill
\subfloat[Manually segmented training binary mask for the above sub-figure (b).
\label{subfig:toads-c}]{%
\includegraphics[width=0.35\textwidth,decodearray={0.5 1}]{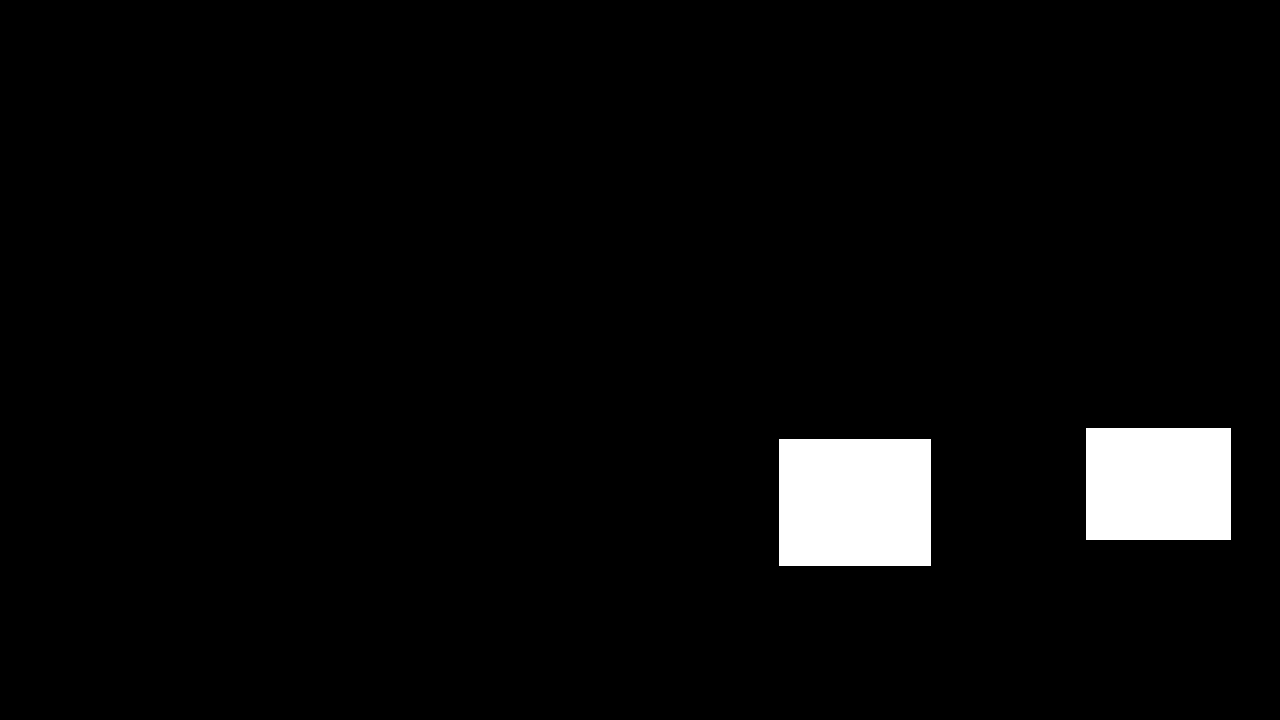}
}
\hfill
\subfloat[Cane toad close-ups. \label{subfig:CaneToads}]{%
\includegraphics[width=0.35\textwidth]{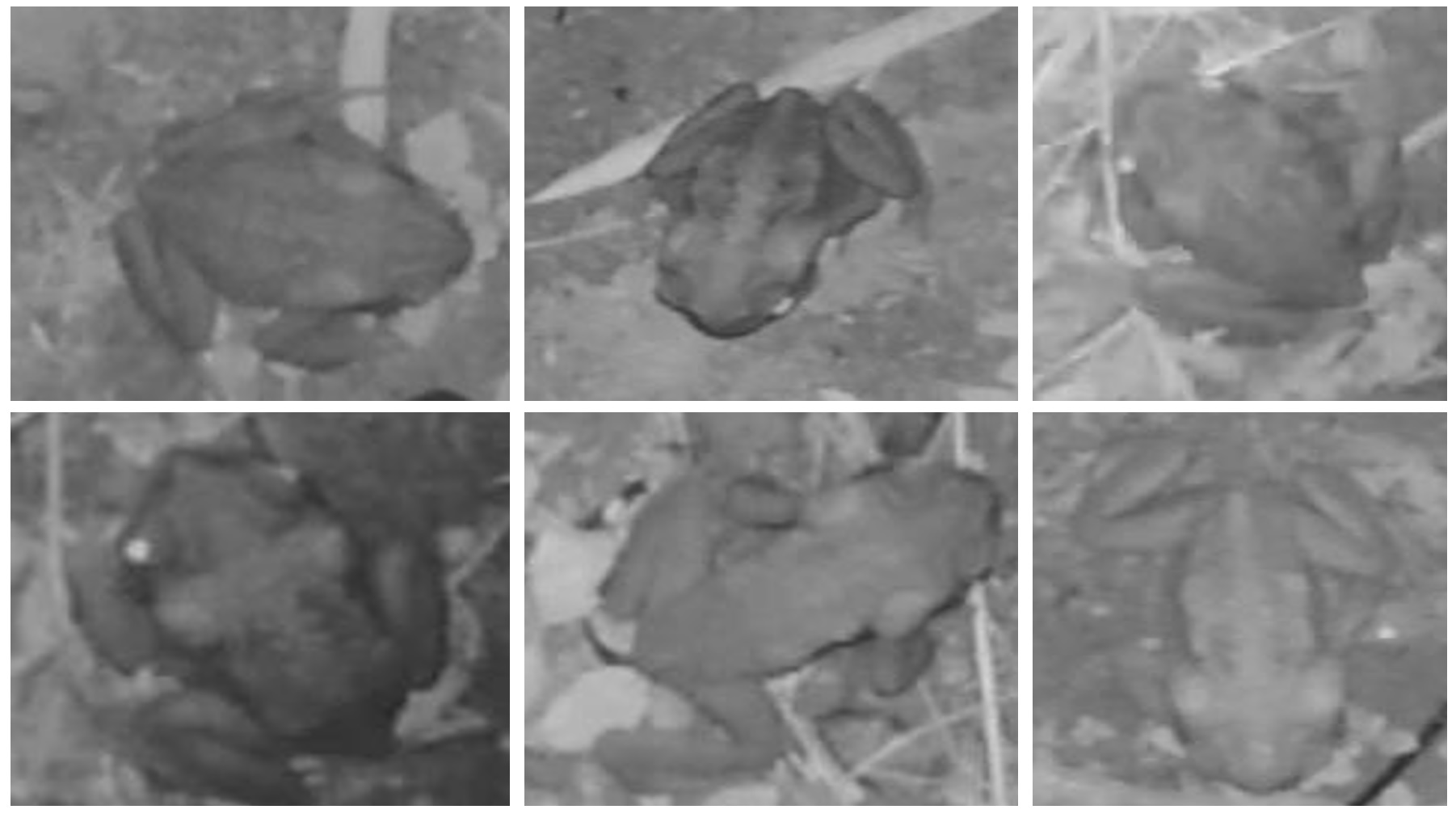}
}
\end{center}
\caption{Typical {\em toad}-labeled video frames on plain and complex backgrounds.}
\label{fig:allToads}
\end{figure}
 
\section{Materials and Methods}
\label{sec:methods}

\subsection{Dataset}
\label{subsec:dataset}

\begin{figure}[!ht]
\begin{center}
\subfloat[Water-holding frogs. \label{subfig:WaterFrogs}]{%
\includegraphics[width=0.35\textwidth]{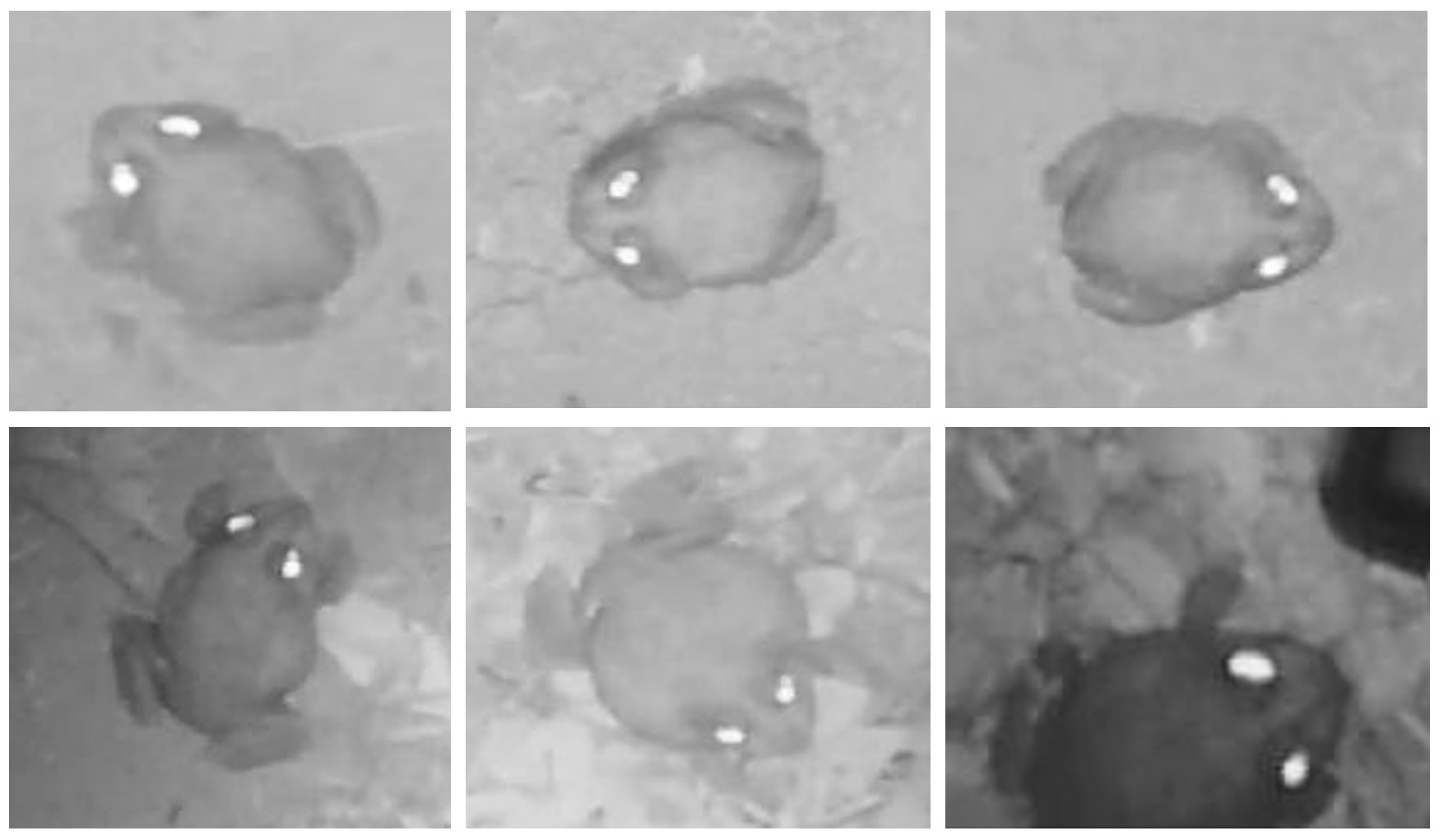}
}
\hfill
\subfloat[Green tree frogs. \label{subfig:GreenFrogs}]{%
\includegraphics[width=0.35\textwidth]{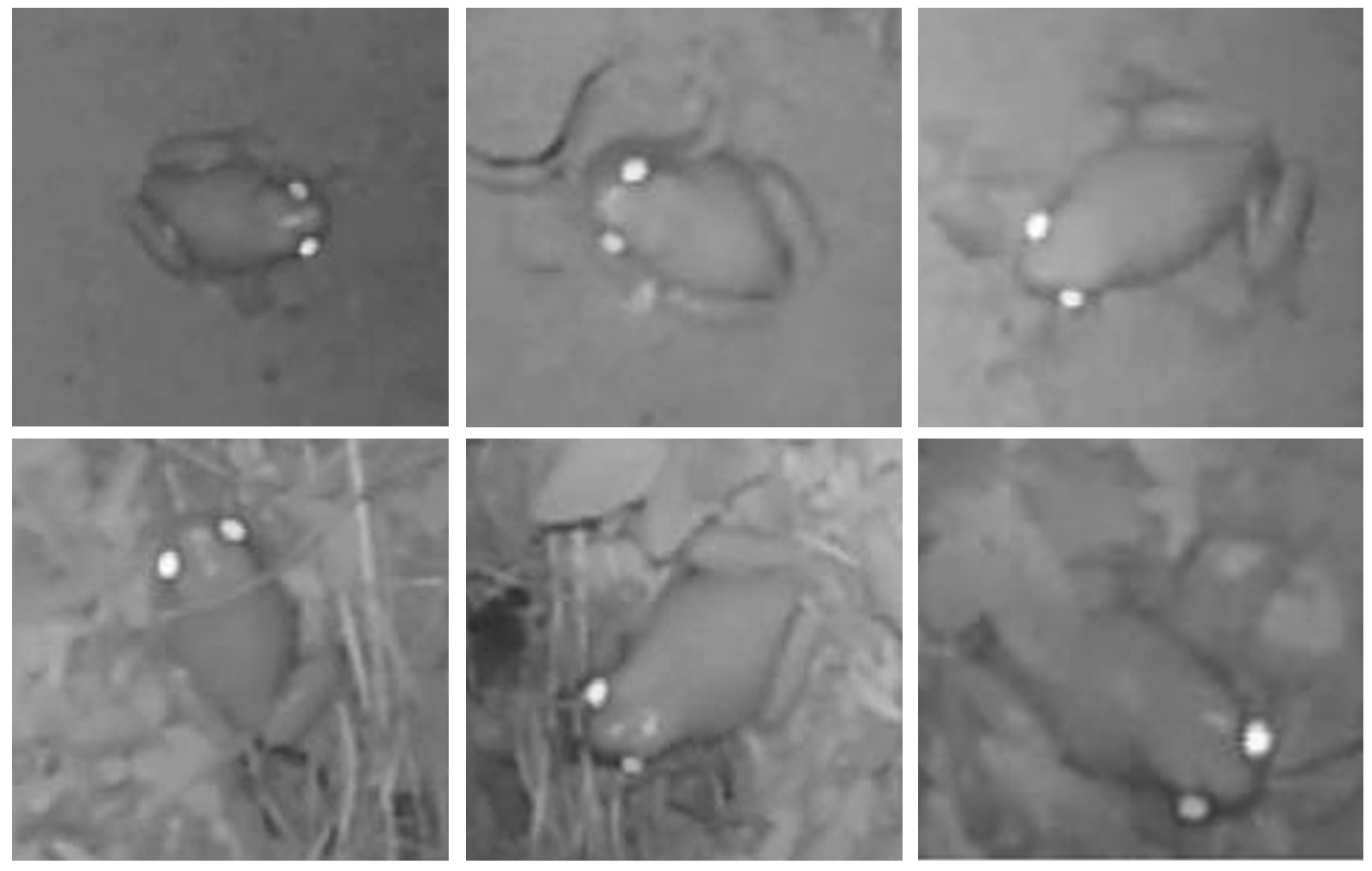}
}
\hfill
\subfloat[Motorbike frogs. \label{subfig:MotorbikeFrogs}]{%
\includegraphics[width=0.35\textwidth]{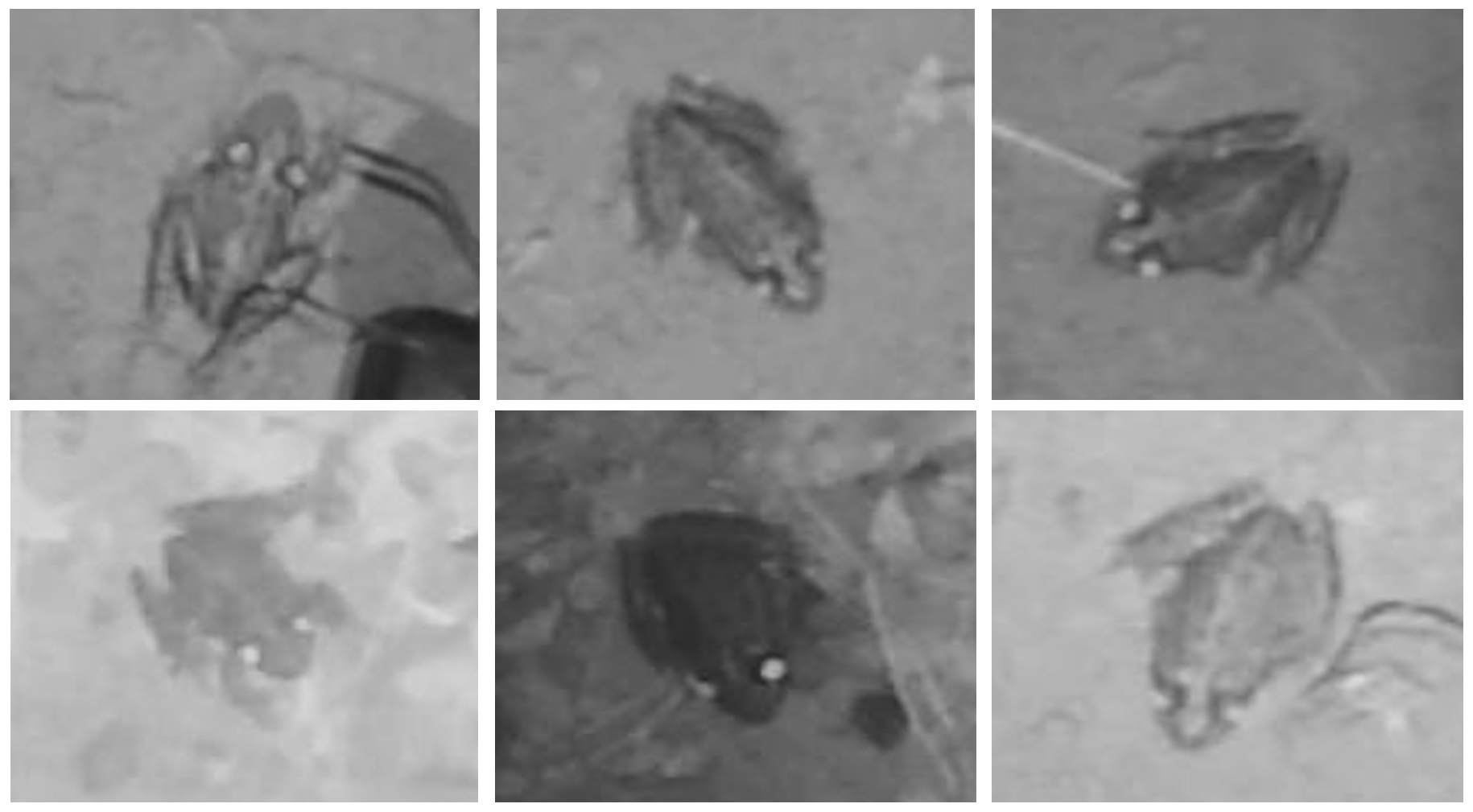}
}
\hfill
\subfloat[Blue-tongue lizards. \label{subfig:Lizards}]{%
\includegraphics[width=0.35\textwidth]{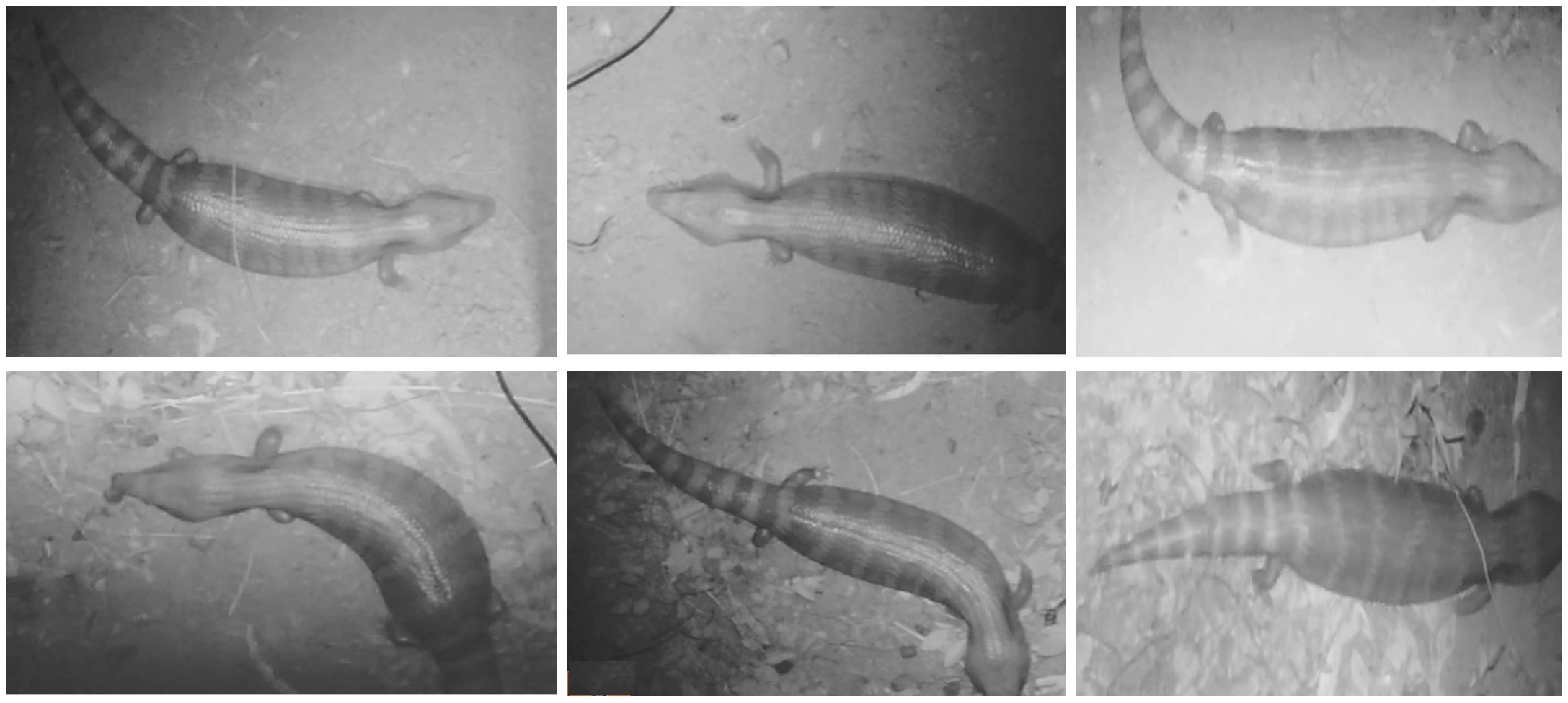}
}
\end{center}
\caption{Typical closeup images of the four native species.}
\label{fig:complex}
\end{figure}

\begin{figure}[!ht]
\begin{center}
\subfloat[Water-holding frog\label{subfig-1:dummy}]{%
\includegraphics[width=0.35\textwidth]{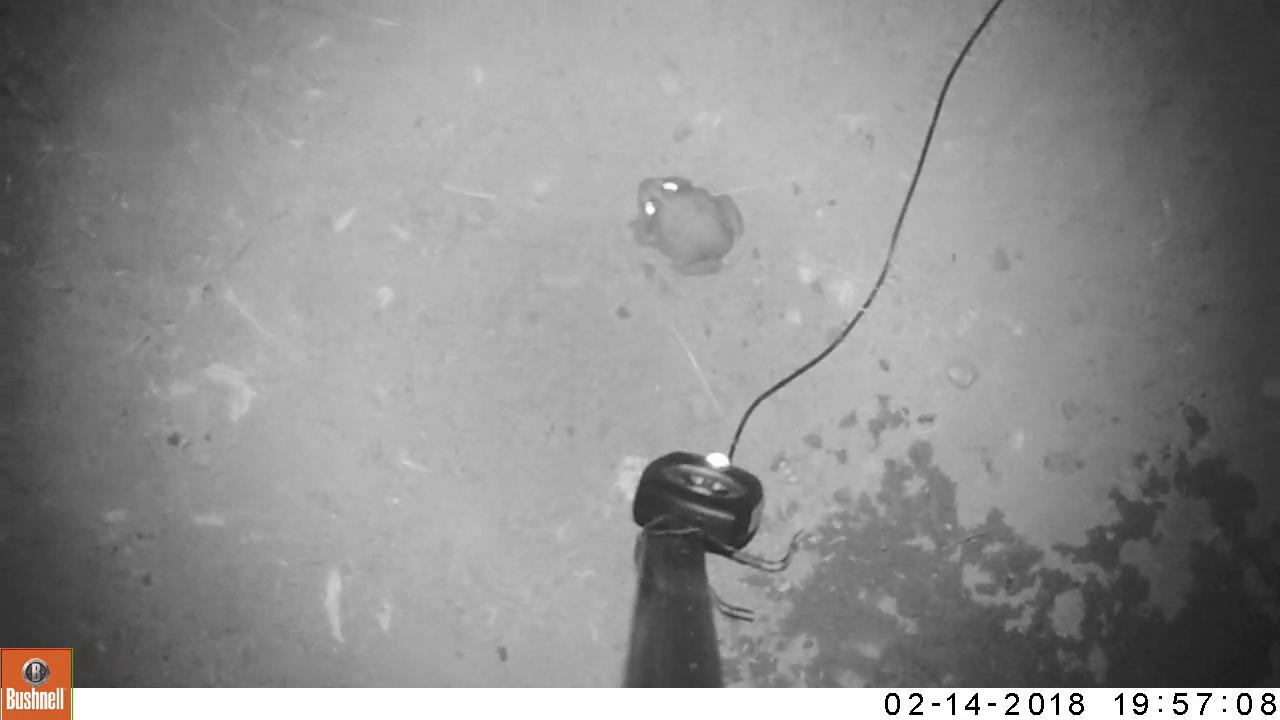}
}
\hfill
\subfloat[Green tree frogs\label{subfig-2:dummy}]{%
\includegraphics[width=0.35\textwidth]{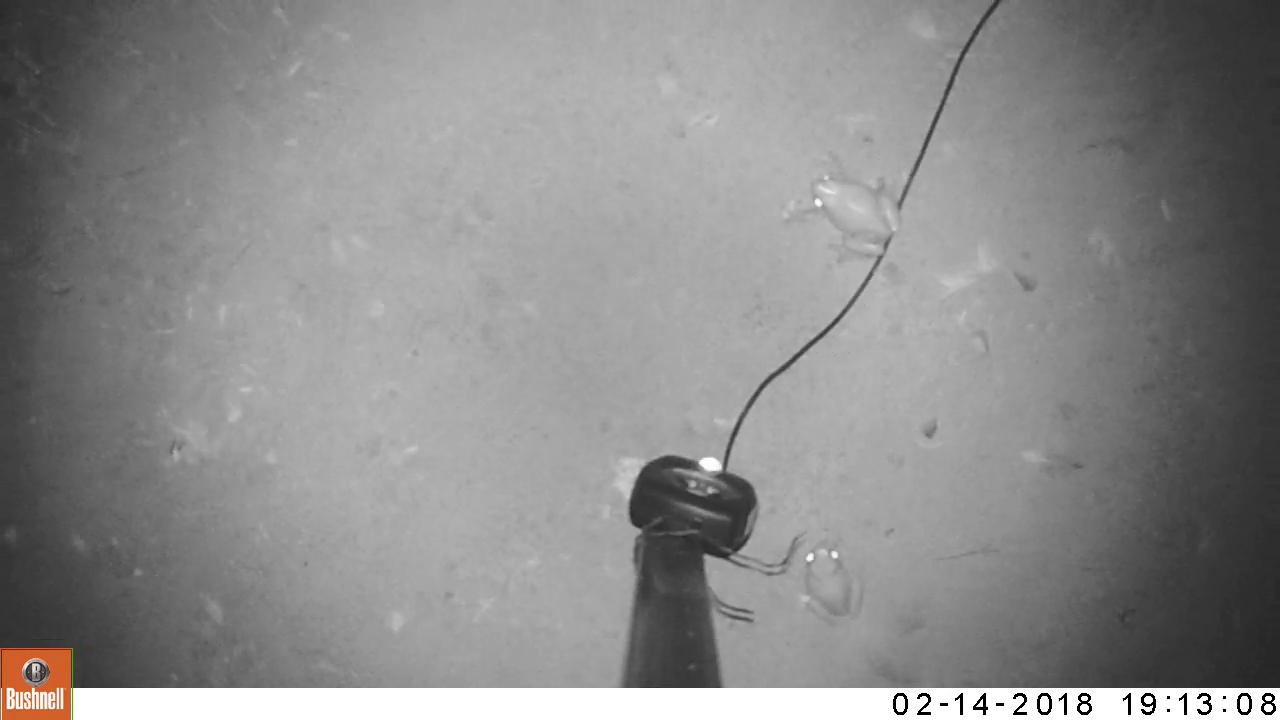}
}
\hfill
\subfloat[Motorbike frog\label{subfig-4:dummy}]{%
\includegraphics[width=0.35\textwidth]{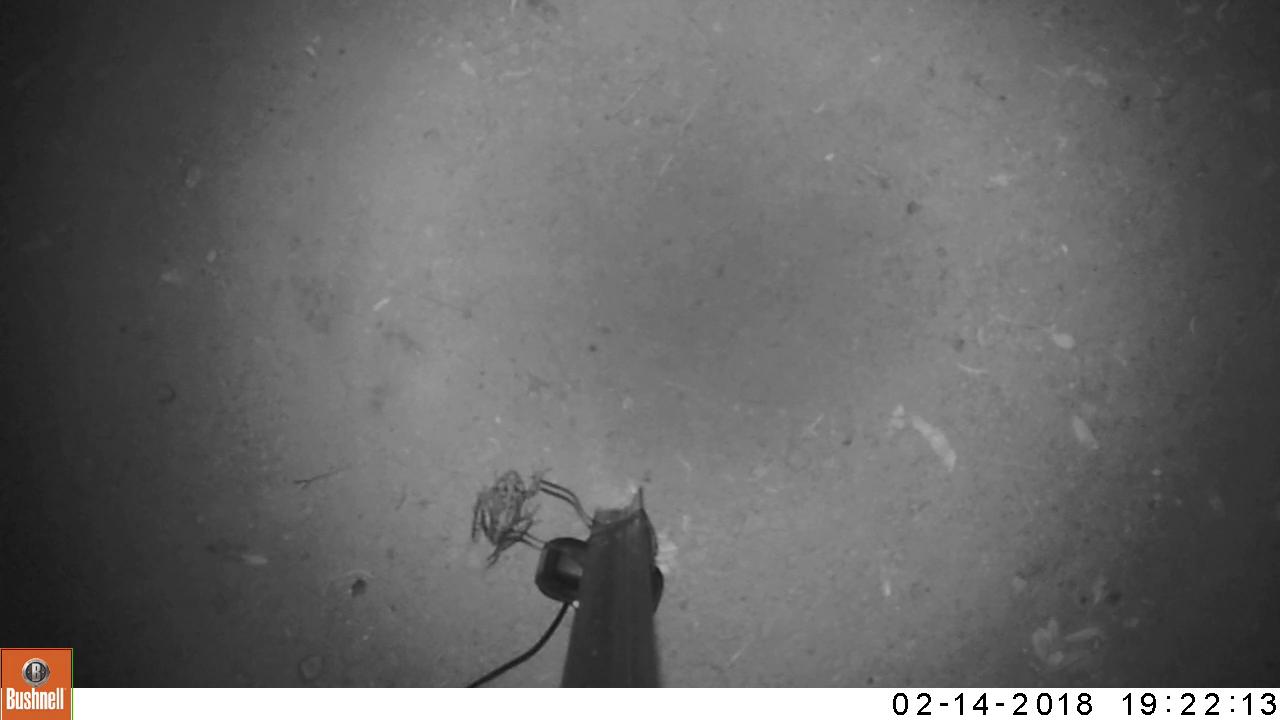}
}
\hfill
\subfloat[Blue-tongue lizard\label{subfig-3:dummy}]{%
\includegraphics[width=0.35\textwidth]{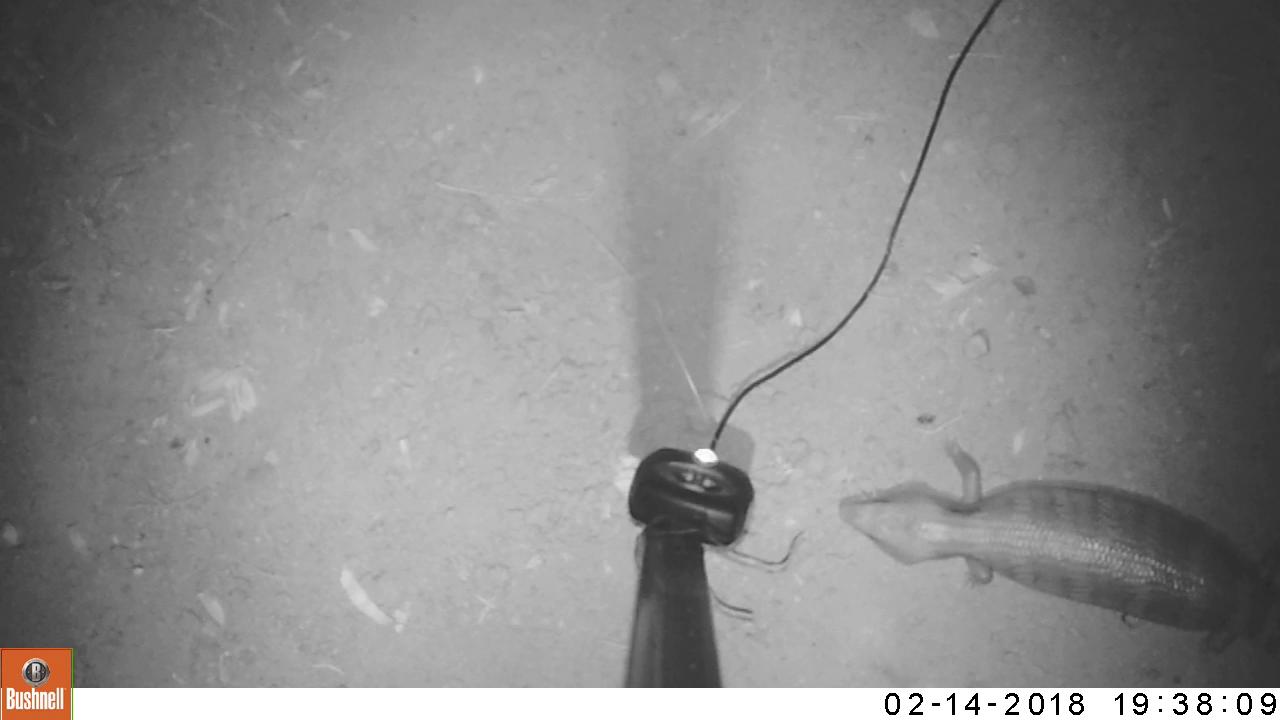}
}
\end{center}
\caption{Typical video frames of four native species on a plain background}
\label{fig:plain}
\end{figure}

Motion activated video cameras were deployed next to a prototype acoustic lure 
\cite{ToadTrap2018} which included an LED ultraviolet light insect lure, and was conceived by the Vertebrate Ecology Lab located at the James Cook University campus in Townsville, Queensland, Australia, with their industry partner Animal Control Technologies Australia. 
Cane toads were identified in 33 and 12 video clips with plain and complex backgrounds, respectively (Fig.~\ref{fig:allToads}). 
Although frogs have not appeared in bycatch for these traps \cite{ToadTrap2018},
native frogs were selected as wildlife to test the visual recognition system because they resemble toads, and could cause confusion for automated recognition. Frog species were selected for their abundance in the local Townsville area, or their resemblance to toads, or both.
The water-holding frog ({\em Cyclorana platycephala}, formerly {\em Litoria platycephala}) was labeled in 20-plain and 4-complex video clips (Fig.~\ref{subfig:WaterFrogs}); 
the green tree frog ({\em Litoria caerulea}) was in 12-plain and 4-complex clips
(Fig.~\ref{subfig:GreenFrogs}); 
the motorbike frog ({\em Litoria moorei}) was in 8-plain and 3-complex clips
(Fig.~\ref{subfig:MotorbikeFrogs});
a blue-tongue lizard ({\em Tiliqua scincoides}) 
was in 9-complex and 4-plain clips 
(Fig.~\ref{subfig:Lizards}).
Each labeled video clip was around 10-20 seconds long and contained only one of the species we examined (Figs~\ref{fig:plain} and \ref{fig:complex}). Total numbers of toad and not-toad video clips were 45 and 64, respectively.

The frames in each of the available video clips were very similar and highly repetitive, and animals were mostly stationary. Therefore, only the first, 42nd, 83rd, etc, frames were extracted (step of 41 frames) from each clip, producing 454 toad and 669 not-toad images. 
The toad-containing images where further examined to select the images with the toads in different locations and/or at different orientations, arriving at 66 distinct images, including the two examples in Fig.~\ref{fig:allToads}.

\begin{figure}[!ht]
\begin{center}
\subfloat[Water-holding frog\label{subfig-1:dummy}]{%
\includegraphics[width=0.35\textwidth]{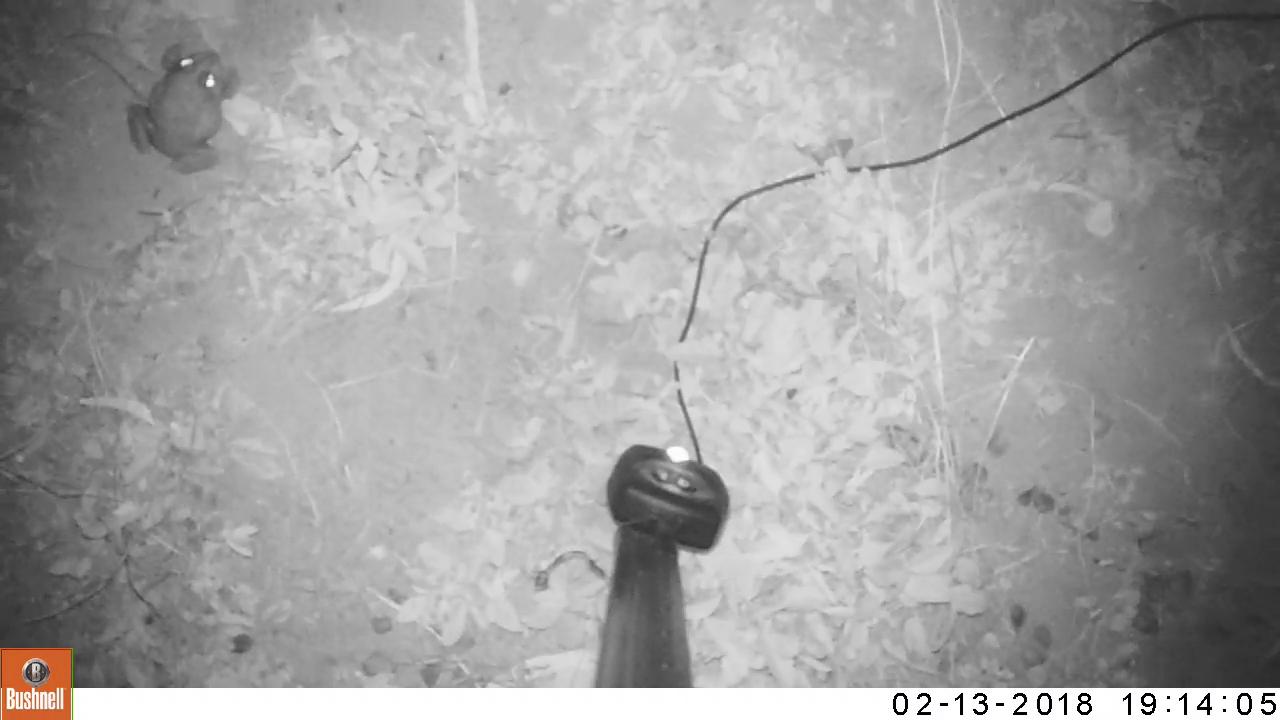}
}
\hfill
\subfloat[Green tree frogs\label{subfig-2:dummy}]{%
\includegraphics[width=0.35\textwidth]{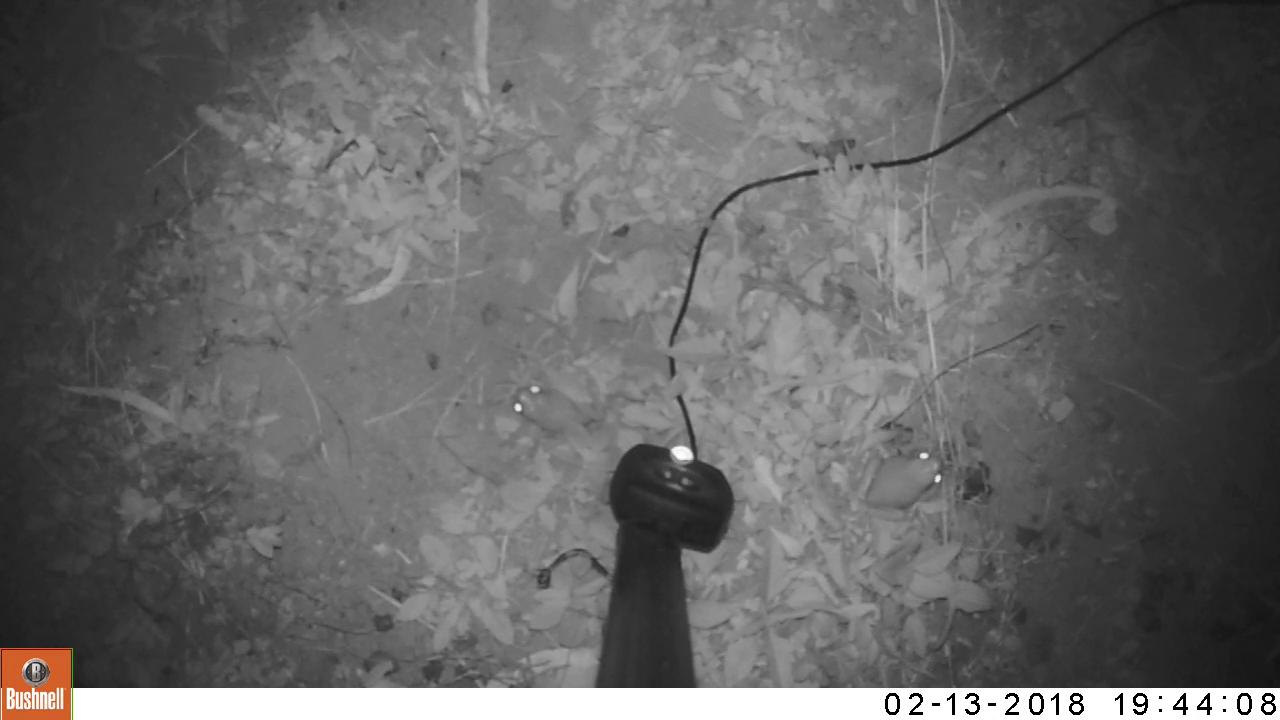}
}
\hfill
\subfloat[Motorbike frog\label{subfig-4:dummy}]{%
\includegraphics[width=0.35\textwidth]{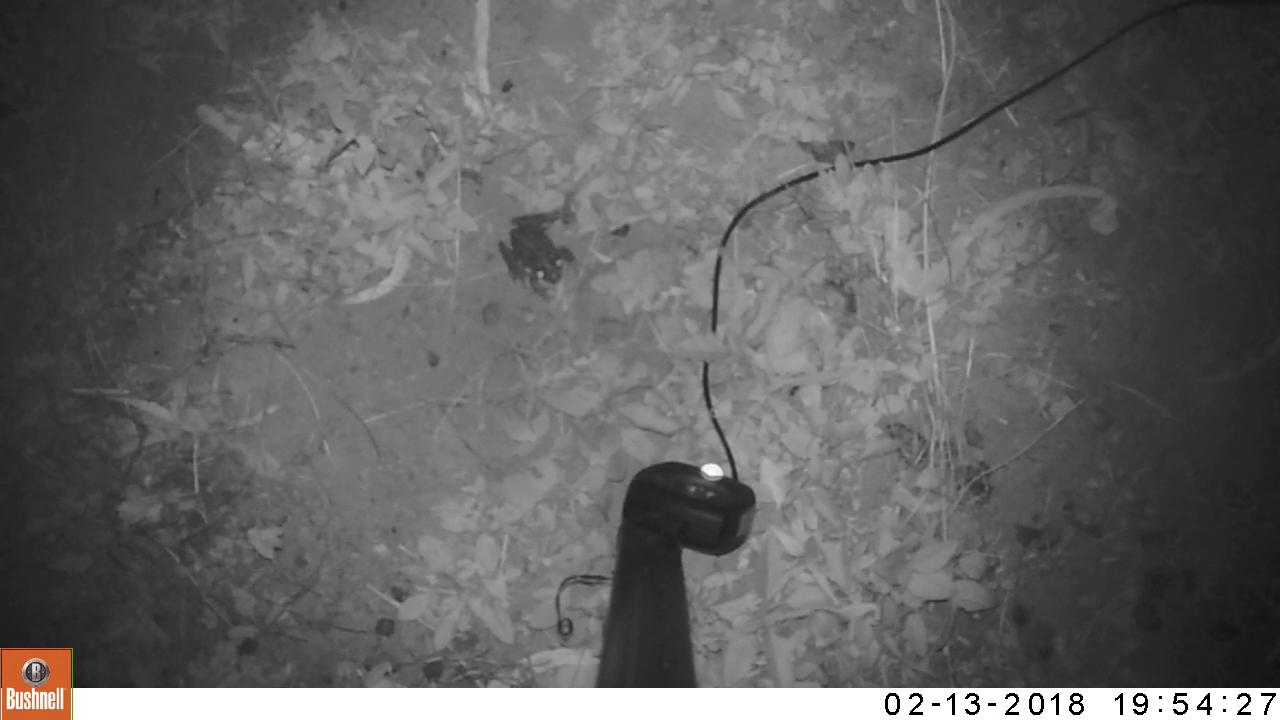}
}
\hfill
\subfloat[Blue-tongue lizard\label{subfig-3:dummy}]{%
\includegraphics[width=0.35\textwidth]{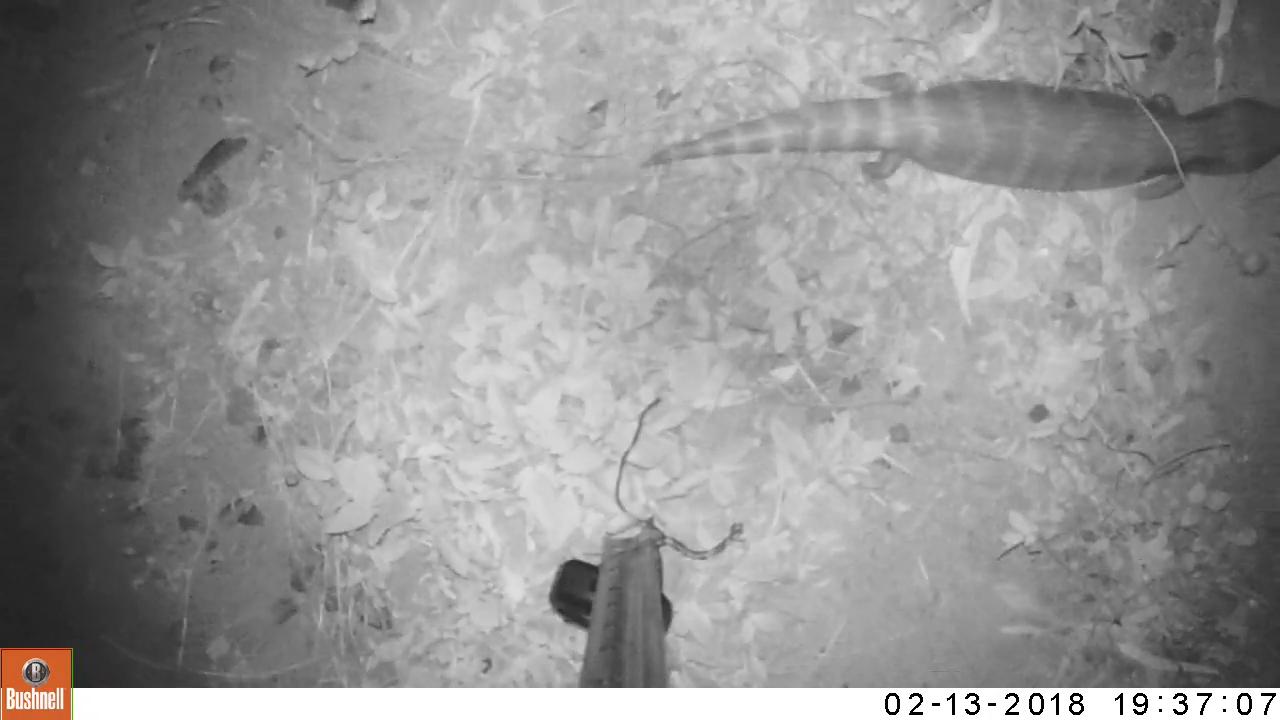}
}
\end{center}
\caption{Typical video frames of four native species on a complex background}
\label{fig:complex}
\end{figure}

\subsection{Detection by heat-map}
\label{subsec:heatmap}

Available in Keras \cite{keras}, the Image-Net-trained Xception CNN \cite{Xception} was selected as the base network using the following reasoning. 
When re-purposed to a single class, Xception contained the smallest number of trainable parameters (20.8 million) compared to 23.5 million in ResNet50 and 21.7 million in InceptionV3. Xception 
is constructed from depthwise separable convolutions, which are growing in acceptance as 
the key building blocks of efficient CNNs \cite{MobileNetV2,NasNet2017}.

The per-image classifier (XToad) was constructed from the Xception CNN, by replacing its 1,000-class top with one spatial average pooling layer followed by a one-class dense layer with a {\em sigmoid} activation function.
Given such a tiny set of 66 training toad images, it became extremely challenging to train XToad without over-fitting on a per-image basis. Therefore, a per-image XToad classifier was not pursued further. 
Note that if a much larger number of training images becomes available, the per-image classifier approach could be a viable option.    

Working within the constraint of the limited number of positive 
(toad-containing) images,
the training capacity of the available 66 images was dramatically enlarged by manually segmenting the cane toads in the images (Fig.~\ref{subfig:toads-c}), where the GNU Image Manipulation Program (GIMP) was used to perform the segmentation. 
The per-pixel XToadHm classifier was then constructed from Xception 
by replacing its 1,000-class top with a $(1 \times 1)$-kernel, single-filter and 
{\em sigmoid}-activated convolution layer, 
i.e. the convolution equivalent of the XToad's dense layer with exactly the same number of trainable parameters (20.8 million). 
The output of XToadHm was a $[0,1]$-ranged {\em heat-map} of an input image spatially-scaled by the factor of 32. For example, if the training images were randomly cropped to $704 \times 704 \times 3$ (from the original $720 \times 1280 \times 3$ shape)
then the XToadHm output was $22 \times 22 \times 1$ tensor of real numbers within 
the $[0,1]$ range. 
To take full advantage of the knowledge-transfer property of the Image-Net-trained Xception, the original three RGB channels were retained, 
but replaced with three identical gray-scale versions of the image. Note that using the identical gray image three times created negligible computational overhead as the $704 \times 704 \times 3$ training image was connected to the first Xception convolution layer with only 864 trainable parameters. 
The weights of the newly created one-class convolution layer were initialized by the uniform random distribution as per \cite{pmlr-v9-glorot10a}, where 
a small regularization weight decay ($1\times 10^{-5}$) was applied to the weights while training. 

\subsection{Training pipeline}
\label{section:training}

The training binary masks were manually segmented as bounding rectangular boxes since the exact outlines of the toads were considered unimportant, 
see an example in Fig.~\ref{subfig:toads-c}. 
Anticipating a much larger number of future training images, 
the bounding boxes were the preferred choice as they could be segmented manually very efficiently for at least a few hundred images. 
For the negative (not-toads) images, the zero-value mask was used. 
All available labeled images, 66 toads and 669 not-toads, were randomly split 80\%-20\%, 
where 80\% of randomly selected images were used as the actual {\em training} subset 
and 20\% were used as the {\em validation} subset, to monitor the training process. 
The random split was controlled by a random seed such that the individual split could be exactly reproduced as required.

While exploring many possible options for training XToadHm, it 
was important to remember that  
the final goal of this project was to deploy the CNN to Internet-of-Things (IoT) devices in the field.
Such IoT devices (e.g., Raspberry Pi) would have limited power and no Internet connection in remote locations where mechanical traps would 
likely be deployed. The goal was, therefore, for XToadHm (or its future equivalents) 
to run on IoT devices and work directly with the 
original $720 \times 1280 \times 3$-shaped images, where any preprocessing should be minimized as it would potentially consume limited battery power.

For training, all images were randomly augmented for each epoch of training, i.e. 
one pass through all available training and validation images. 
Specifically, the Python bindings for OpenCV (Open Source Computer Vision Library) package were used to perform the following 
augmentations, and in the following specified order, where each 
image, and, if applicable, the corresponding binary mask were:
\begin{enumerate}
\item randomly cropped $720 \times 720$ from the original $720 \times 1280$ pixels, 
where a $rows \times columns$ convention was used throughout this work to denote the spatial dimensions;
\item randomly spatially rotated in the range of  $[-360, +360]$ degrees;
\item randomly shrunk vertically in the scale range of $[0.9, 1]$ and independently horizontally within the same range. 
More severe proportional distortions could potentially confuse the cane toads (Fig.~\ref{subfig:CaneToads})
with the water-holding frogs (more rounded, Fig.~\ref{subfig:WaterFrogs}) 
and/or motorbike frogs (more elongated, Fig.~\ref{subfig:MotorbikeFrogs});
\item transformed via random perspective transformation to simulate a large variety of viewing distances and angles; 
\item flipped horizontally with the probability of 50\%;
\item randomly cropped  to retain the final training $704 \times 704 \times 3$ input tensor $X$, and the corresponding $704 \times 704 \times 1$ target mask tensor $Y$;
\item divided by 125.5 for the $[0,255]$-ranged color channel values, i.e. as per the intended use of Xception \cite{keras};
\item mean value subtracted;
\item randomly scaled intensity-wise by $[0.75,1.25]$;
\end{enumerate}
If the image $I$ of $[0,255]$-range values was the result of the above steps 1-6 then the steps 7-9 converted $I$ into 
the training tensor $X$ via 
\begin{equation}
X=(Z - mean[Z])\times s, \ Z=I/125.5,  \ s \in [0.75,1.25].
\label{eq:X_meanX}
\end{equation}
Note that the original Xception was trained with $X=Z - 1$ instead of the step number 8. Equation (\ref{eq:X_meanX}) removed the ability of the CNN to 
distinguish toad/not-toads using the image intensity, where in testing, $s=1$ was used.
After the steps 1-6, the augmented target mask $Y$ was downsized to a $22 \times 22$ shape to match the XToadHm output. Without the preceding extensive augmentation, XToadHm easily over-fitted the available training images, by essentially memorizing them, 
i.e. achieving very low training loss values without the corresponding reduction of the validation loss.

The standard per-pixel binary cross-entropy (\ref{eq:L}) was considered first with $W_t=1$, 
\begin{equation}
loss = -W_t y \log (p) - (1-y) \log(1-p),
\label{eq:L}
\end{equation}
where $y$ was the given {\em ground truth} pixel mask value, 
and $p$ was the per-pixel output of the XToadHm network. 
There were many more negative images, 669, than positive images, 66. 
Furthermore, the total nonzero area was much smaller than the zero area in the toad-containing positive masks. Due to such a significant imbalance of negative and positive training heat-map pixels, the non-weighted loss ($W_t=1$) collapsed the CNN output to near-zero values. 
Thus, the toad-class weight $W_t$ was set to 100 by the order-of-magnitude estimation of the ratio of negative to positive training pixels.
The weight value ($W_t=100$) was not optimized further to avoid over-fitting the available 
training images.

The Keras implementation of Adam \cite{Adamax14} was used as the training optimizer. The Adam's initial learning-rate ($lr$) was set to $lr=1\times 10^{-4}$, where the rate was halved every time the total epoch {\em validation} loss did not decrease after 10 epochs. The training was done in batches of 4 images (limited by the GPU memory) 
and was aborted if the validation loss did not decrease after 32 epochs, where the validation loss was calculated from the validation subset of images. 
While training, the model with the smallest running validation loss 
was saved continuously. 
If the training was aborted, the training was restarted four more times 
but with the previous starting learning rate halved, i.e. $lr=0.5 \times 10^{-4}$ for the first restart, $lr=0.25 \times 10^{-4}$ for the second restart, etc.
Each restart loaded 
the previously saved model with the smallest validation loss achieved so far. 
Note that not only the training images but also the validation images were augmented by the preceding augmentation pre-processing steps in order to prevent indirect over-fitting of the validation images.
It took approximately 6-8 hours to train an instance of XToadHm on Nvidia GTX 1080Ti GPU.

\subsection{Training by Gaussian heat-maps}
\label{subsec:gauss}

After extensive experiments with the preceding training pipeline, it became apparent that  large residual training and validation losses were due to
inherited segmentation errors, 
i.e. the cane toads were deliberately segmented 
only approximately by bounding rectangular boxes.
Since the precise toad contours were not required, 
the hand-segmented binary boxes were converted 
to 2D Gaussian distributions \cite{Bulat2017,NIPS2014_5573} via 
\begin{equation}
Y(r,c) = \exp \big( -(r-\bar{r})^2/a  - (c-\bar{c})^2/b \big),
\label{eq:gaussY}
\end{equation}
\begin{equation}
\bar{r} = (r_{\min} + r_{\max})/2, \ \ \ \bar{c} = (c_{\min} + c_{\max})/2,
\label{eq:mean_r}
\end{equation}
where: $r$ and $c$ were the pixel row and column indexes, respectively;
the minimum and maximum toad-bounding-box row values were $r_{\min}$ and $r_{\max}$, and 
similar for the columns $c_{\min}$ and $c_{\max}$;
and where $a$ and $b$ constants were determined from 
\begin{equation}
-(r_{\min}-\bar{r})^2/a  = - (c_{\min}-\bar{c})^2/b  = \ln(0.5)
\label{eq:GaussDrop}
\end{equation}
on a per-box basis. 
The $0.5$ constant in (\ref{eq:GaussDrop}) was the reduction ratio 
of the Gaussian amplitude from its center (value of one) to the box boundaries.

\begin{figure}[!ht]
\begin{center}
\includegraphics[width=0.49\textwidth]
{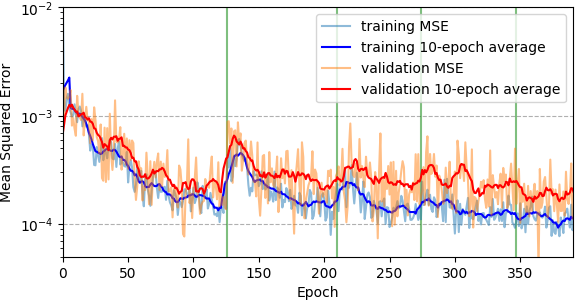}
\end{center}
\caption{Training and validation MSE losses from the XToadHm CNN training history.}
\label{fig:training}
\end{figure}

By converting the sharp-edged binary toad-bounding-boxes to 
2D Gaussian distributions mean-centered at the box's geometrical centers, the problem of toad image segmentation was transformed into a toad localization problem. The use of 2D Gaussians in a localization problem is a powerful technique currently used in the more complex problem of human pose estimation \cite{Bulat2017,NIPS2014_5573}, 
where even occluded human body landmarks need to be localized in the images. Because the training target masks became non-binary, the Mean Squared Error (MSE) was used as the training loss instead of the binary cross entropy (\ref{eq:L}). Furthermore, and somewhat surprisingly, the MSE loss did not require the class-balancing weight $W_t$ (\ref{eq:L}) to handle the highly imbalanced numbers of positive and negative training pixels. The actual (and very typical) training history for the final version of XToadHm CNN is shown in 
Fig.~\ref{fig:training}, where vertical lines indicate the restarted training with learning rate halved. The training MSE loss was trailed very closely by the validation MSE loss 
(Fig.~\ref{fig:training}) indicating negligible over-fitting issues.

\begin{figure}[!ht]
\begin{center}
\includegraphics[width=0.49\textwidth]
{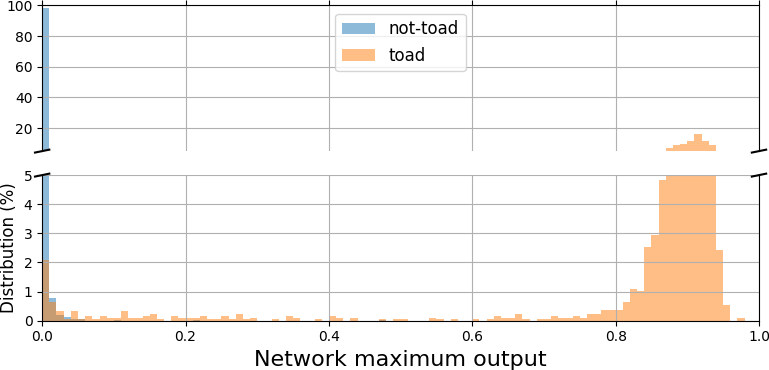}
\end{center}
\caption{Normalized histograms of XToadGmp outputs in 0.01 steps for the test {\em not-toad} and {\em toad} images.}
\label{fig:histogram}
\end{figure}

\begin{table}[!ht]
\renewcommand{\arraystretch}{1.3}
	\caption{Confusion matrix 
    \label{table:confusion}}
	\centering
    \begin{tabular}{ c | c | c }
		\hline
		\hline
          & {\bf Actual Positives} & {\bf Actual Negatives}  \\
          &  {\bf {\em Toad}} &  {\bf {\em Not-toad}}  \\
		\hline
      {\bf Predicted Positives} & $TP=1728$ & $FP=0$   \\
    {\bf Predicted   Negatives} & $FN=135$ & $TN=2892$  \\
  		\hline
        {\bf Column totals} & $P=1863$ & $N=2892$  \\
  		\hline
  		\hline
	\end{tabular}
\end{table}

\begin{table}[!ht]
\renewcommand{\arraystretch}{1.3}
	\caption{Performance metrics 
    \label{table:metrics}}
	\centering
    \begin{tabular}{ c | c }
		\hline
		\hline
             {\bf {\em Recall}} & $TP / P = {\bf 92.7\%}$    \\
             {\bf {\em Precision}} & $TP / (TP + FP) = {\bf 100\%}$    \\
             {\bf {\em Accuracy}} & $(TP + TN) / (P + N) = {\bf 97.1\%}$     \\
             {\bf {\em F-measure}} & 
             $2 / (1/precision + 1/recall) = {\bf 96.2\%}$  \\
  		\hline
  		\hline
	\end{tabular}
\end{table}

\begin{figure}[!ht]
\begin{center}
\subfloat[Fragment of the XToadHm heat-map output spatially-enlarged 
and multiplied by the corresponding {\em toad}-containing input image. 
\label{subfig:HeatmapToads}]{%
\includegraphics[width=0.35\textwidth]{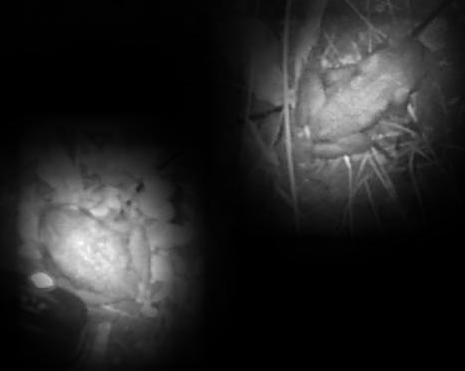}
}
\hfill
\subfloat[Example of false-negative: predicted as {\em not-toad}, where the cane toad was jumping (top center-right). \label{subfig:JumpingToad}]{%
\includegraphics[width=0.35\textwidth]{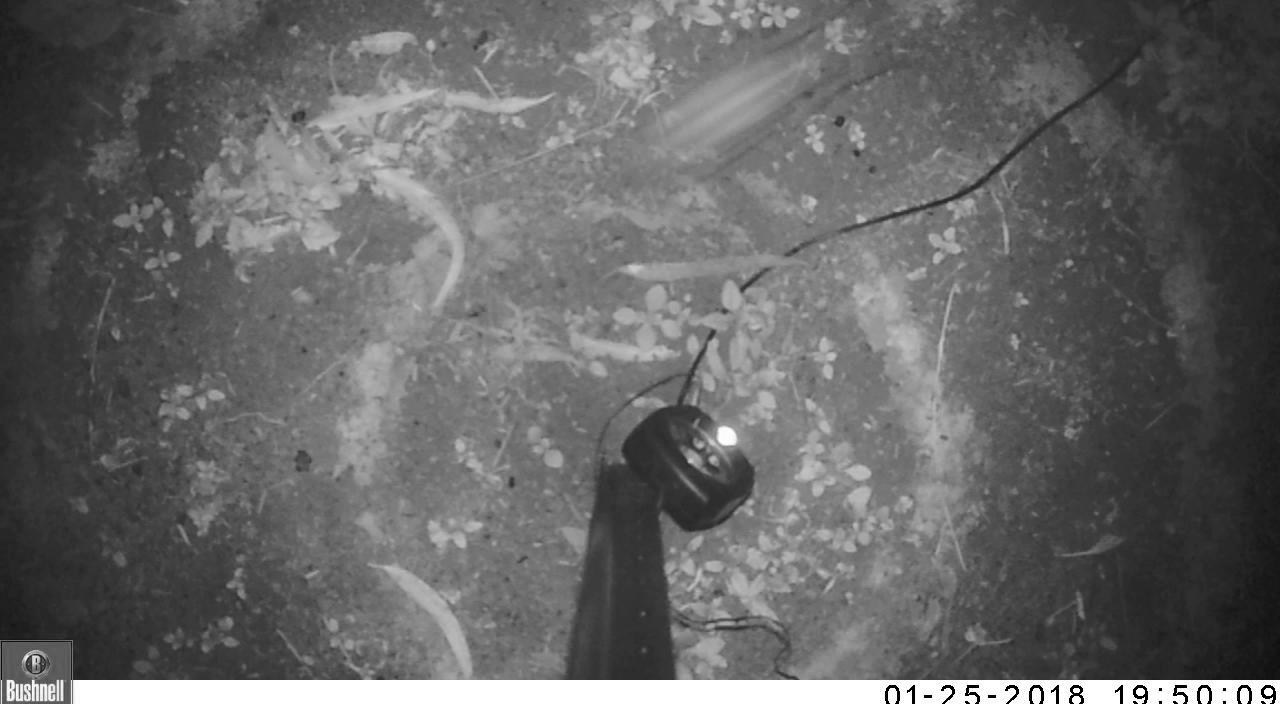}
}
\hfill
\subfloat[A mislabeled {\em toad} frame predicted as {\em not-toad}. 
\label{subfig:mislabledFrame}]{%
\includegraphics[width=0.35\textwidth]{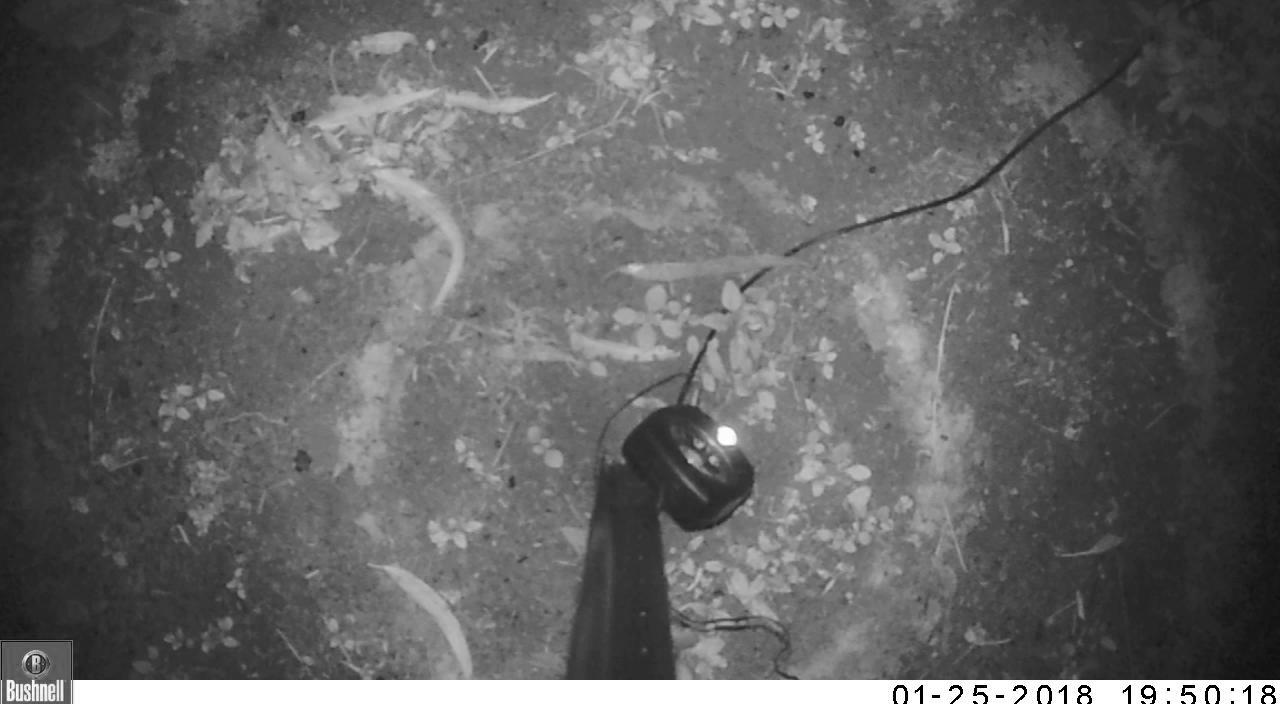}
}
\end{center}
\caption{Prediction examples.}
\label{fig:toads}
\end{figure}

\section{Results and Discussion}
\label{sec:results}

Keeping in mind the final desired deployment of the Cane Toad Vision Detection system 
onto low-power, low-cost IoT devices, 
the prediction version of the XToadHm CNN should not use any additional computationally expensive post-processing of its sigmoid-activated heat-map output.
Thus, prediction XToadGmp CNN was constructed by appending the global spatial maximum pooling layer (hence the {\em Gmp} abbreviation) to the output of the XToadHm CNN.
The XToadHm CNN was trained (Fig.~\ref{fig:training}) on the $704 \times 704$ shaped images. 
Due to its fully convolutional nature, 
the trained XToadHm CNN could be re-built into 
XToadGmp CNN to accept any image shape, where $704 \times 1280$ input shape was used for testing.

The {\em test} images were extracted from the available labeled videos with the step of 9 frames starting from the 10th frame, which made all test images different from the {\em training} and {\em validation} images.  For prediction, the test images were 
$704 \times 1280$-center-cropped (from the original $720 \times 1280$), converted to the gray-scale for each color channel, divided by 125.5 and mean-subtracted 
(see steps 7 and 8 in Section~\ref{section:training}). 

When the XToadGmp CNN was applied to the test images, it produced the $[0,1]$-range outputs, which exhibited very wide amplitude separation between the {\em not-toad} (near-zero outputs) and {\em toad} (larger than 0.5 outputs) test images, see Fig.~\ref{fig:histogram}. The {\em toad}-detection threshold was left at the default 0.5 value.
The confusion matrix (Table~\ref{table:confusion}) and common performance metrics were used \cite{ROC06} and summarized in Table~\ref{table:metrics},
where actual {\em vs.} predicted instances were denoted as total actual positives ($P$), 
total actual negatives ($N$), predicted true-positives ($TP$), 
true-negatives ($TN$), false-positives ($FP$) 
and false-negatives ($FN$, number of actual toads predicted as not-toads). 

The XToadGmp CNN achieved 0\% {\em false-positive} rate (Table~\ref{table:confusion}), which was highly desirable CNN property in order to avoid trapping native species. 
The heat-map XToadHm CNN (rebuilt for $704 \times 1280$-inputs) 
was also applied to the test images to confirm that the heat-map
outputs were remarkably accurate in locating the cane toads, even when there was more than one toad in the image, see example in Fig.~\ref{subfig:HeatmapToads}.
Misclassified as {\em false-negative} test images were reported and some of them were examined. Examination revealed that  
misclassification occurred in some instances when the cane toad was jumping, 
see example in Fig.~\ref{subfig:JumpingToad}. Since, however, the XToadGmp CNN correctly detected the toads in the frames before and after the jump, such transitional frames were
not flagged as an issue and were left in the reporting results. 
All false-negative images were examined, and many images without cane toads
were found, see example in Fig.~\ref{subfig:mislabledFrame}. 
Such clearly mislabeled video frames were removed from the results. Most of the remaining false-negatives contained partially visible cane toads, e.g. occluded by the central box or located at the edges of the images.

\section{Conclusion}

In conclusion, this study developed a novel approach for training an accurate
Convolutional Neural Network image-classifier 
from a very limited number of positive images 
using Gaussian heat-maps, where only 66 toad-containing images were 
used.
The Image-Net-trained Xception CNN \cite{Xception} was end-to-end re-trained by the new approach and achieved 0\% {\em false-positives}, 
92.7\% {\em recall}, 
100\% {\em precision}, 
97.1\% {\em accuracy}, 
and 96.2\% {\em F-measure} ({\em f1-score}) on the 4,755 {\em in-situ} test images (Tables~\ref{table:confusion} and \ref{table:metrics}), which were not used in training.

\bibliographystyle{IEEEtran}
\bibliography{IEEEabrv,toads2018}
%
 
\end{document}